By Robert Griffin, Tyson Cobb, Travis Craig, Mark Daniel, Nick van Dijk, Jeremy Gines, Koen Krämer, Shriya Shah, Olger Siebinga, Jesper Smith, and Peter Neuhaus

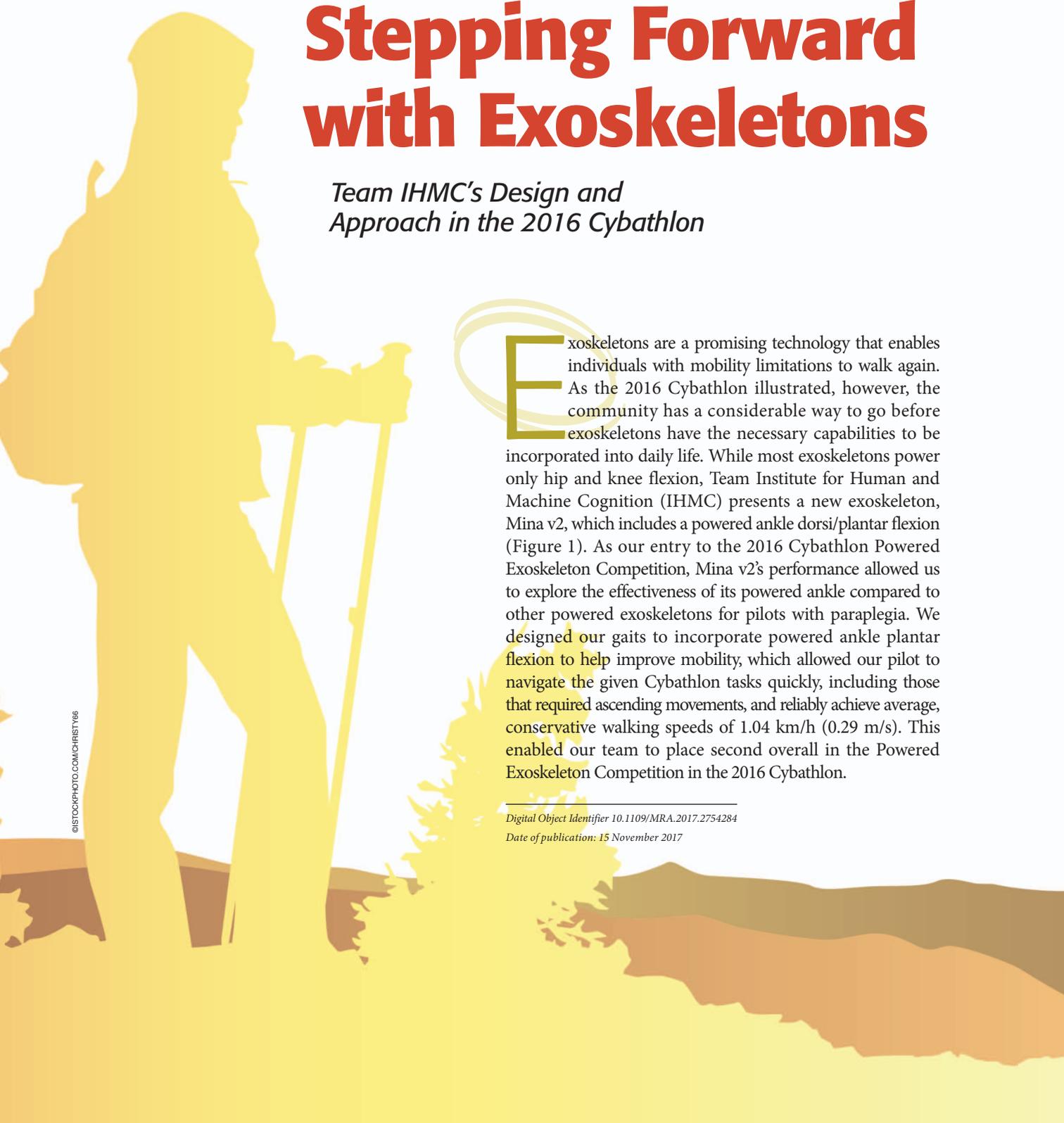

# Stepping Forward with Exoskeletons

*Team IHMC's Design and Approach in the 2016 Cybathlon*

Exoskeletons are a promising technology that enables individuals with mobility limitations to walk again. As the 2016 Cybathlon illustrated, however, the community has a considerable way to go before exoskeletons have the necessary capabilities to be incorporated into daily life. While most exoskeletons power only hip and knee flexion, Team Institute for Human and Machine Cognition (IHMC) presents a new exoskeleton, Mina v2, which includes a powered ankle dorsi/plantar flexion (Figure 1). As our entry to the 2016 Cybathlon Powered Exoskeleton Competition, Mina v2's performance allowed us to explore the effectiveness of its powered ankle compared to other powered exoskeletons for pilots with paraplegia. We designed our gaits to incorporate powered ankle plantar flexion to help improve mobility, which allowed our pilot to navigate the given Cybathlon tasks quickly, including those that required ascending movements, and reliably achieve average, conservative walking speeds of 1.04 km/h (0.29 m/s). This enabled our team to place second overall in the Powered Exoskeleton Competition in the 2016 Cybathlon.

## Exoskeleton Background

Exoskeletons, long an idea of science fiction, have the potential to change day-to-day life for countless individuals, particularly those with mobility issues. While approximately 70% of people with spinal cord injury paraplegia use a manual wheel-chair [1], being seated for extended periods causes a variety of other medical concerns, e.g., degradation of bone density [2], muscle atrophy [3], and pressure sores [4], in addition to requiring special infrastructure adaptations such as ramps and lifts to conduct daily life activities. To address this, commercial exoskeletons such as ReWalk [5], Ekso [6], and Indego [7] have made significant strides forward since the first exoskeleton prototypes. Indeed, the recent 2016 Cybathlon illustrated the incredible progress that exoskeletons have made in recent years, with many now capable of ascending and descending stairs and ramps. However, while the Cybathlon included some tasks of daily living, they were presented in an idealized situation, far from what one would encounter in the real world; realistically speaking, handrails aren't always available, steps are of different heights, and surrounding crowds interfere with crutch placement. While several of the pilots were able to complete most of the challenges, the Cybathlon thoroughly demonstrated how far exoskeletons have to go before pilots are able to walk with the speed and ease of an able-bodied person.

A variety of exoskeletons have been developed for improving mobility. The ReWalk was one of the first such devices to show the potential for restoring ambulation for those limited to a wheelchair. Three exoskeletons have been approved by the U.S. Food and Drug Administration for use as rehabilitation devices on flat ground: ReWalk [5], Ekso [6], and Indego [7]. Each of these devices has demonstrated sit-to-stand capabilities, although ReWalk is leading the way by providing users the ability to ascend and descend stairs, as well.

All three platforms feature motors at the hips and knees to power the leg motions. Each executes a position trajectory at the command of the pilot, triggered by either a body tilt or button press, to perform the walking motion. The ReWalk and Ekso can also prematurely stop this stepping motion if ground contact is detected. Additionally, the ReWalk can command different step types through a wrist-mounted interface. While walking and standing, however, all of the balance and stability is provided by the pilot through the use of forearm crutches. Overall, the mobility capabilities of even these most advanced devices are limited to rehabilitation centers or home use.

We previously developed the Mina v1 [8] and NASA X1 exoskeletons [9] (the latter in collaboration with NASA). Similar to the three commercial devices, these two exoskeletons feature actuators at the hips and knees: harmonic drive reduced dc motors for Mina v1 [8] and custom rotary series elastic actuators on X1. Series elastic actuators were also used by the University of Twente, The Netherlands in the design of lower-extremity powered exoskeleton (LOPES), a gait-training exoskeleton that operates by setting joint impedances for the hips and knees using custom Bowden-cable series elastic actuators to adjust the user's gait [10]. For purposes of rehabilitation, controlling the impedance allows the user to adjust the amount of assistance the device provides, leading to potentially effective therapies. Additionally, impedance control is now the standard actuation approach for legged robotics, as it enables compliant interactions with the environment. For mobility assistance, impedance control thus offers the potential for powered exoskeletons to provide locomotion on par with humanoid robots.

Despite trying to restore upright sagittal plane mobility, powered exoskeletons have differed from their biological counterparts in one critical way: powered exoskeletons typically lack ankle actuation. Able-bodied walking relies on the motion and forces exerted by ankle plantar flexion and dorsiflexion. During flat walking, approximately 40–50% of positive power is provided by the ankle joint [11]. It has been found, however, that as much as 40% of the positive work done by the entire leg during walking comes from energy stored in the ankle muscle–tendon system [11]. Despite this, powered ankle plantar flexion is rare in powered exoskeletons. Indeed, the most common powered exoskeletons (ReWalk [5], Ekso [6], and Indego [7]) do not have powered ankles but instead use a passive, spring-loaded joint.

While there have been many powered exoskeletons that include powered ankles, these are typically ankle-only exoskeletons for data collection [11] or specific orthotic purposes [12] or, alternately, augmentation exoskeletons [13]. To explore the effects of ankle plantar flexion and dorsiflexion on exoskeleton systems, our new exoskeleton, Mina v2 (pictured in Figure 2), includes powered hip flexion/extension, knee flexion, and, notably, powered ankle dorsi/plantar flexion.

The Powered Exoskeleton Competition in the Cybathlon consisted of a variety of tasks, including a slalom course, ascending and descending a ramp that is not compliant with the Americans with Disabilities Act (ADA), navigating a tilted path, ascending and descending a set of stairs, and traversing a set of stepping stones, in addition to sitting down and standing up from a seat. In this article, we present the strategies that we developed for accomplishing these tasks. Our approach uses powered ankle flexion to reduce pilot effort as much as possible by including powered toe-off motions in the trajectories. Unlike many other exoskeleton gait designs, ours provides a walking gait utilizing a transfer phase that includes a large degree of toe-off. To increase the flexibility of the walking gait, we also developed a unique method for finding joint-position trajectories based on defining Cartesian waypoints for the swing foot. We believe that the use of toe-off motion combined

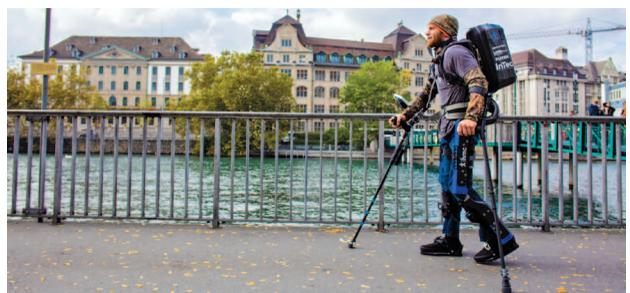

**Figure 1.** IHMC's pilot, Mark Daniel, walking with the Mina v2 exoskeleton in Zürich, Switzerland.

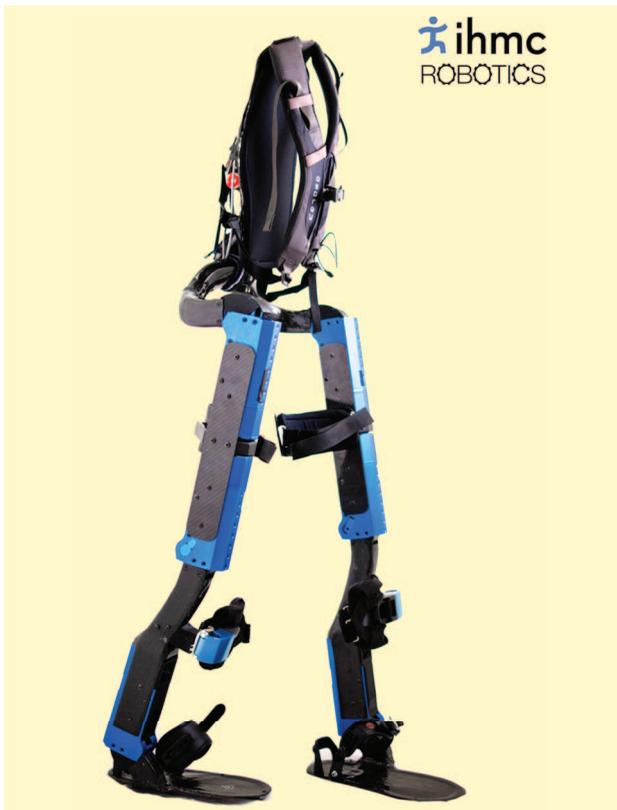

**Figure 2.** An image of the Mina v2 exoskeleton, shown without the backpack.

with our flexible trajectory design was critical to our team's success in the 2016 Cybathlon.

### Design of Mina v2

Designed as our entry to the 2016 Cybathlon, Mina v2 is the latest exoskeleton developed by IHMC. This design drew on our experience with the design and manufacture of Mina v1 [8], the NASA X1 exoskeleton [9], and the Hopper exercise exoskeleton [14]. A complementary work describing the hardware in more detail is being prepared.

### Mechanical Design

Mina v2 features a fully custom carbon-composite design. The device includes six electric actuators, integrated into the structure as load-bearing components, and a protective backpack for the electronics. Mina v2 functions as a prototype device, designed and built to custom dimensions specifically to fit our pilot. Future modifications will include adjustable links to fit other pilots, the design of which were not feasible within the time constraints of this project.

The actuators themselves are custom linear linkage actuators that are modular in construction to allow for ease of replacement, accessibility, and repair. They were designed in-house, specifically for use with Mina v2, and feature a frameless electric motor, integrated electronics, and an onboard motor amplifier and controller for distributed joint-level control. The structure of the actuator uses a slider-crank linkage mechanism driven by a linear ball-screw transmission. The slider-crank mechanics are tuned and optimized to produce the highest torque and gear ratio for the desired range of motion. In the current application, each linear linkage actuator produces approximately 110-Nm peak torque and no-load velocity of 9 rad/s about the point of rotation, with a 130° range of motion, powered by an Allied Motion HS02303 motor with a continuous and peak current of 6 A and 20 A, respectively.

To house the actuators, each link of the device is a custom-manufactured composite piece. The link structure is first designed and 3D-printed as an ABS plastic mold. The mold is used to maintain internal tolerances while layers of carbon fiber weave and unidirectional cloth are laid over it. Epoxy resin is then vacuum-infused into the cloth. Once the epoxy has cured, the joint structures are demolded and post-processed to fit. This utilization of carbon fiber components keeps the total exoskeleton mass, including the power electronics and backpack, to 34 kg.

### Electrical Design

Each actuator is equipped with a magnetic incremental encoder on the motor, a magnetic absolute encoder on the output, and a load cell on the output of the linkage. The motor is controlled by a Twitter Gold motor drive (Elmo Motion Control, Petach-Tikva, Israel) capable of performing position, velocity, and current control, depending on the desired control mode. The motor drive is mounted on a custom carrier board that breaks out the connections for the encoders and the load cell.

All of the other electrical components are housed in the 7.5-kg backpack. Central control is performed on an embedded COM Express Type 6 computer (ADLINK Technology, Inc., New Taipei City, Taiwan) running a custom Ubuntu kernel. The control code runs on a Java virtual machine using POSIX real-time threads using a custom library. The embedded computer communicates with the motor drivers over EtherCAT. The EtherCAT line is split into two separate lines by an Omron EtherCAT junction, allowing for more efficient wiring of the legs. This Omron junction also allows the connection of other EtherCAT-enabled sensors, such as an inertial measurement unit.

Mina v2 is powered by a 48-V, 480-Wh lithium ion battery (designed for electric bicycles) and is capable of approximately 2.5 h of fully powered autonomous runtime. The battery has an onboard battery-management system to protect from current overdraw and under-voltage conditions. A wireless emergency stop and secondary wired emergency stop are included to interrupt motor power in case of an error. Watchdog timers monitor communications with the motor controllers and can disable the controller if needed.

### Trajectory Design

Mina v2 was designed to explore the effects of including powered ankle plantar flexion on an orthotic robotic exoskeleton. As such, the trajectories are designed around the use of this additional degree of freedom. The powered ankle plantar flexion allows the trailing leg to apply greater forces to the

68 • IEEE ROBOTICS & AUTOMATION MAGAZINE • DECEMBER 2017

ground when walking and performing other functions. This section outlines the details for calculating the desired trajectories, exploiting the described toe-off action.

### Design of Swing Leg Trajectory

To increase the flexibility of the walking gait, we designed the swing-leg trajectory in task space, making it a function of Cartesian waypoints. We defined four Cartesian waypoints with corresponding spatial velocities, as illustrated in Figure 3: the starting and desired ending locations (the red dots) and two middle waypoints (the blue dots). The two midpoints are defined at a tuneable percent of the step length, $\%l_{s,f}$, $\%l_{s,b}$, and a desired swing height, $h_s$, in Figure 3. These parameters are then selected, for both normal walking and stair ascension, heuristically through testing. We then use inverse kinematics to solve for the necessary hip and knee angles and velocities at the waypoints that begin each step, and we set these as the boundary conditions for minimum-jerk joint trajectories. This allows the step parameters to be changed online and incorporated at the next step. Stance-leg trajectories are additionally generated to straighten the knee and rotate at the hip and ankle to move the pilot forward during the step, as illustrated by the gray leg silhouette in Figure 3.

This approach for calculating swing-leg joint angles was used for all the tasks in the Cybathlon, from walking on flat ground to stairs to slopes. By defining the trajectories in this fashion, changing step lengths and times did not require additional effort beyond changing the final goal position and the total trajectory time. Not only did this ease the development effort when the pilot was learning to use the exoskeleton, but also it accelerated the training process, as all tuning could be performed online rather than through code changes. Additionally, only one code module was required to calculate joint trajectories, rather than separate code for each different Cybathlon task.

### Design of Transfer Trajectory

In natural, able-bodied walking, ankle plantar flexion is used to inject energy into the system, starting at the end of the swing phase and continuing through transfer [15]. To more closely emulate this, we introduce a toe-off motion during a short transfer phase at the beginning of every step. This motion consists of commanding a minimum-jerk trajectory that ends at a certain angle to the trailing ankle during transfer. This corresponds to a change in the leading-leg hip flexion as the body rotates about the leading ankle [see Figure 4(a) and (b)]. An additional, fast toe-off motion is added at the beginning of the swing phase to impart an additional impulse to the system, as shown in Figure 4(c).

While the powered ankle does not necessarily enable dynamic walking equivalent to that of an able-bodied individual, it does provide several benefits. Energy injected by the powered ankle effectively reduces the amount of additional "pushing" with the upper body required by the pilot during walking, potentially decreasing the overall required exertion. While there are ways to equivalently inject this energy with-out a powered ankle, such as first bending the leg and then quickly straightening it, these are undesirable as they move further away from natural walking gaits. Moreover, a powered toe-off motion reduces the required crutch force by moving the pilot kinematically into a more advantageous position to both begin and to continue walking. We plan to analyze these effects in more detail in future work.

This toe-off motion worked quite well, helping the pilot when continuously walking and providing considerable assistance when starting from rest. This is a common situation, as the lack of hip internal/external rotation requires a "skid-steer" style of ambulation that requires frequent pauses to reposition. In the Cybathlon, the stepping-stone task, in particular, was aided by the powered toe-off motion, as the need for precise foot placement required stopping between each step, resulting in a long stance length that needed considerable effort on the part of the pilot to resume walking.

### Design of Stairs Trajectory

To make stair ascension easier for our pilot, we designed joint trajectories to capitalize on the powered ankle plantar flexion. Uniquely, our approach did not take steps one at a time, as other exoskeletons typically do, instead stepping with only one foot on each step.

Our approach for stair ascension is outlined in Figure 5. Between steps, the hips are positioned evenly between the feet and then moved directly over the leading ankle before stepping up, as shown in Figure 5(a) and (b). The goal is to move the center of mass directly over the leading ankle, making balancing easier for the pilot, and also to have all of the actuator's work go into raising the center of mass. This motion is combined with significant toe-off push, so that the center of mass will begin to be raised during transfer, when both legs are on the ground. Including this motion greatly decreases the required torque at the upcoming support knee. This, in turn, results in lessening the pilot's effort. Similar to our toe-off motion during flat walking, an additional fast toe-off motion at the end of the transfer phase adds an impulse to the system, as shown in Figure 5(c).

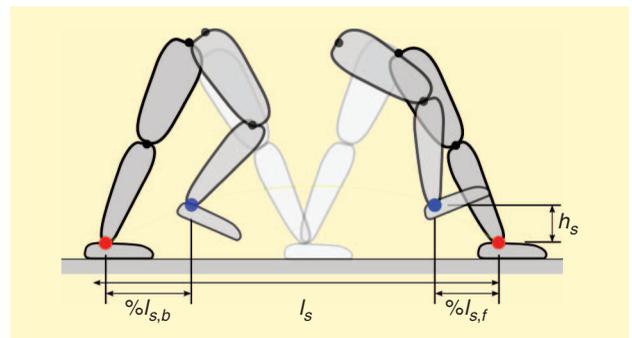

**Figure 3.** An illustration of the swing trajectory plan. The calculated midpoints and endpoints are shown by the blue and red dots, respectively. The stance leg is shown by the light gray silhouette. The necessary joint positions and velocities are computed at each of the waypoints.

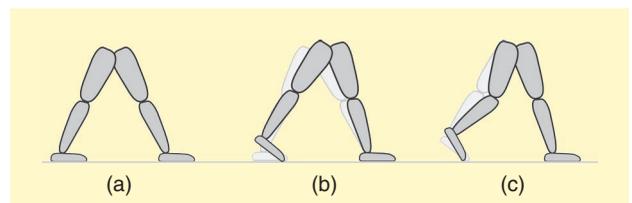

**Figure 4.** An illustration of the toe-off motion used during transfer with a time-lapse shown by (a)–(c).

Instead of descending the stairs in the traditional forward manner, we chose to descend backwards, similar to how a person descends a ladder. We believed that this would provide greater stability and control to our pilot, making it much less likely for him to fall down the stairs. We followed the same approach as for ascension, using only one foot for each step, without, however, including toe-off; as the pilot is descending, this is unnecessary. This approach is highlighted in Figure 5 (f)–(i).

### Design of Ramp Trajectory

Similar to the other tasks, the trajectories we designed for ascending and descending a ramp revolve around the use of powered ankle plantar flexion. For ascending, we use the same approach as in flat walking, except that the hips are only slightly in front of the trailing foot at the start of transfer. For descending, we similarly place the hips over or only slightly in front of the trailing foot. Also, there is no additional fast toe-off motion at the end of transfer when descending. This approach is illustrated in Figure 6.

### Pilot Interface

The human–machine interface for Mina v2 is based on manual input by the pilot. It consists of a Raspberry Pi 3 computer with a screen, a thumb joystick, and a momentary switch, all of which are mounted on the right crutch. The joystick is integrated in the front-facing part of the handle, and the switch is mounted as a trigger on the bottom of the handle. The human–machine interface runs off a separate battery pack and communicates with the embedded computer over transmission control protocol/Internet protocol (TCP/IP). The connection can be made using WiFi, making the whole crutch wireless. However, during the competition, a wired connection was used to avoid interference.

The joystick is employed to change behaviors; the legality of these behavior changes was tailored to the competition. The trigger button acts as a play cue and will initiate movement; every step is triggered separately. This gives the pilot the ability to synchronize his weight shift and the start of a step himself but also allows for a brief recovery after an unbalanced step. The next step can be triggered 0.25 s before the end of the current step, allowing the pilot a continuous walking motion if desired.

### Results and Discussion

In preparation for the Cybathlon, extensive training was undertaken so that our pilot could complete tasks in as little time as possible. Prior to this event, our pilot had approximately 20 h of experience in previous exoskeletons over the course of six years. Our pilot was confident enough with the walking gait that, after the competition, he used Mina v2 to navigate the streets of Zürich, as shown in Figure 1. For the competition, the low-level motor controllers were set to track desired positions, but also had the capability of controlling the joint impedance, which is being investigated for future work as we explore the dynamic effects of our toe-off motion in greater depth.

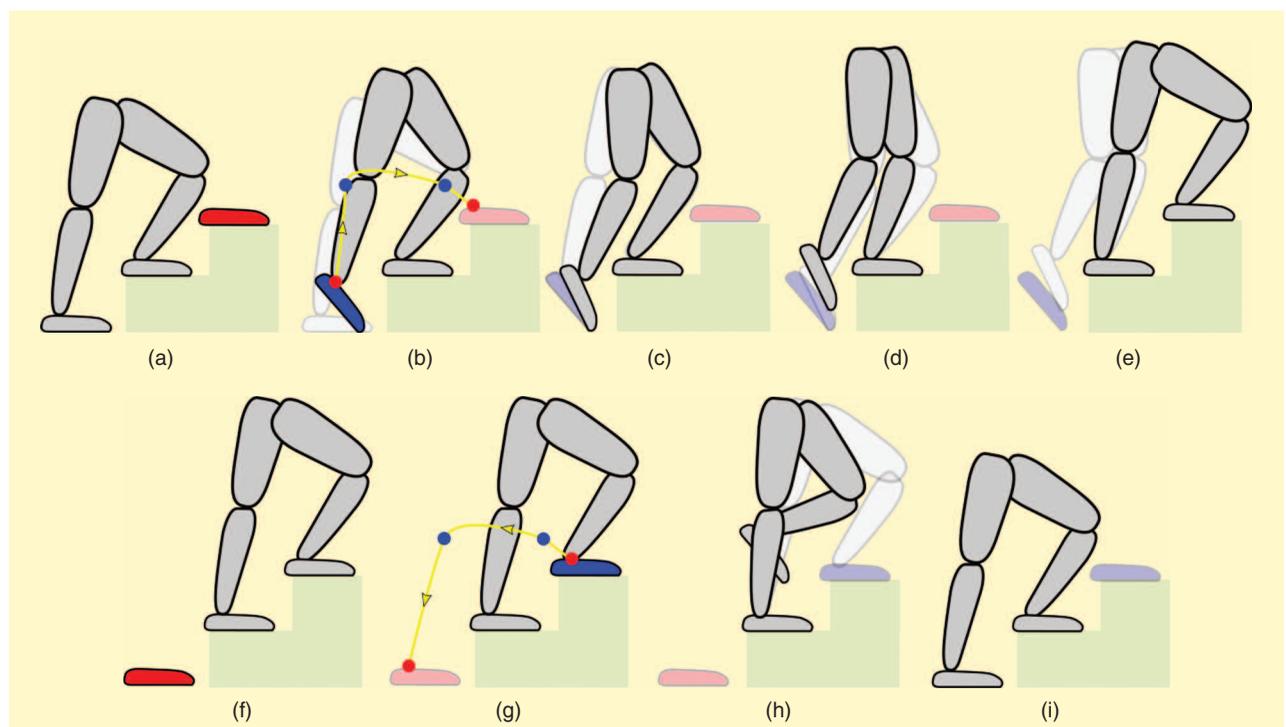

**Figure 5.** An illustration of stair ascent (a)–(e) and descent (f)–(i) with powered ankles.

### Inclusion/Exclusion Criteria for Cybathlon

We required the pilot to have an American Spinal Injury Association Impairment Scale rating of either A (Complete) or B (Incomplete). Before each session, the pilot was examined for preexisting bruises, open sores, skin lesions, or skin irritations. After each day of training, the pilot was examined to determine if the use of Mina v2 caused any bruising, chafing, or skin irritations.

### Trajectory Design Results

Table 1 lists the parameters used to determine the trajectories for the swing and stance legs using our approach. These parameters were selected heuristically through testing to generate appropriate trajectories that are executable by the pilot. Factors such as feedforward motion, pilot comfort, and pilot stability are examined throughout the tuning process.

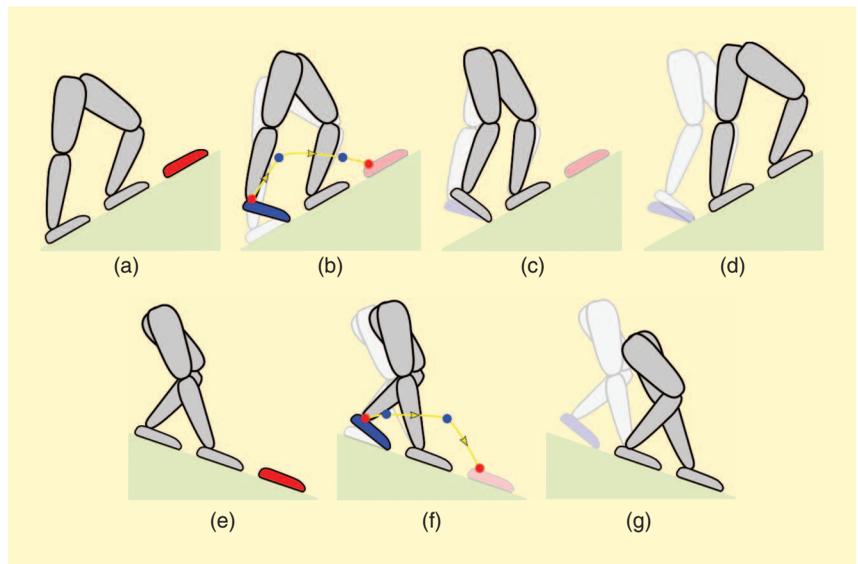

**Figure 6.** An illustration of ramp ascent (a)–(d) and descent (e)–(g) with powered ankles.

**Table 1. A Table of the Parameters Used to Generate the Swing- and Stance-Leg Trajectories.**

|  | Flat Ground | Stairs | Slopes | Stepping Stones |
|---|---|---|---|---|
| $l_s$(m) | 0.4 | 0.29 | 0.31 | $\geq 0.35, \leq 0.69$ |
| $h_s$(m) | 0.1 | 0.15 | 0.08 | 0.1 |
| $\%l_{s,b}$ | 15 | 20 | 20 | 15 |
| $\%l_{s,f}$ | 15 | 20 | 20 | 15 |
| Swing time (s) | 1.0 | 1.6 | 1.2 | 1.8 |
| Transfer time (s) | 0.4 | 1.1 | 0.6 | 0.6 |

### Walking

When walking over flat ground, the desired step length is set to 0.4 m, with a transfer duration of 0.4 s and a swing duration of 1.0 s, resulting in an average velocity of 1.04 km/h (0.29 m/s). While this results in the transfer phase making up a considerably higher percentage of the gait than in able-bodied walking, it provides the pilot adequate time to comfortably and reliably reposition the crutches between steps when walking continuously. Both transfer and swing durations could be reduced through more training, as well as increasing step length. However, these parameters produced a fairly comfortable gait for the pilot during training. He was able to maintain this walking speed for a significant duration without substantial fatigue, often training for over an hour at a time. Higher walking speeds were attempted, but performance was not reliable enough for the competition, given the relatively brief training period.

The resulting trajectories for the transfer and swing leg walking on flat ground are illustrated in Figure 7, which shows the average swing and stance leg-joint angles during a step. The duration has been normalized to the step time. The shaded gray region represents the transfer phase of the walking gait. The resulting trajectories are smooth. There is a significant toe-off motion during transfer, with the change in the stance-hip angle moving the pilot forward, which helps the pilot start walking.

### Stepping Stones

The stepping stones task at the Cybathlon consisted of a series of small wooden platforms that the pilot had to step on sequentially, without allowing his feet to touch the ground around the platform. This required long steps, typically starting from a static position. A large toe-off motion was utilized during transfer to help propel the pilot forward off of his trailing foot and toward the next stone. We found that being able to do so greatly assisted our pilot in starting his step to the next stone. This task, in particular, highlights the benefits of including a powered toe-off motion in the walking gait, as we were one of only two teams able to successfully complete the task during the competition. Figure 8 illustrates how this large toe-off angle helps the pilot move his weight toward his stance foot before swing.

### Stairs

The stair height for the Cybathlon was 18 cm, and we used a transfer duration of 1.1 s and swing duration of 1.6 s, resulting in the joint trajectories shown in Figure 9. While this is substantially slower than what is used in flat walking, it allowed our pilot to shift his weight forward over the leading foot before raising it. Unlike other exoskeletons, which typically only step up with one side, our pilot ascended a stair with each step, as Figure 5 illustrates, similar to typical able-bodied motion. We believe this was made possible by the use of powered ankle plantar flexion. Our pilot was able to utilize this approach quite successfully in practice and in the competition, reliably and quickly ascending the stairs. These trajectories were then played backwards, without the toe-off motion, to descend the stairs.

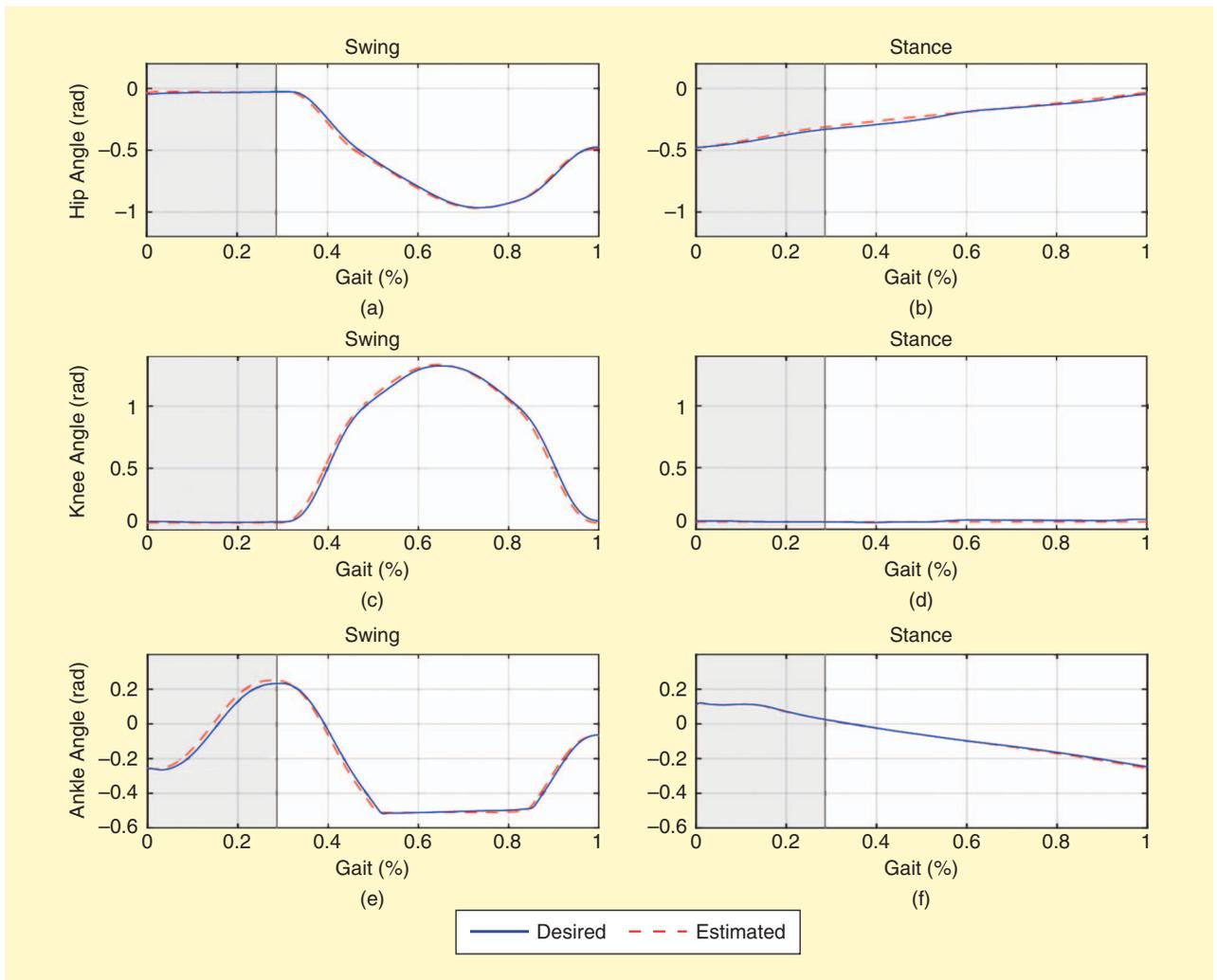

**Figure 7.** The joint angles when walking on flat ground with powered ankles; the shaded gray region represents the double-support state: hip angle during (a) swing and (b) stance; knee angle during (c) swing and (d) stance; and ankle angle during (e) swing and (f) stance.

### Cybathlon Results

With our hardware and trajectory design, we placed second in the 2016 Cybathlon Powered Exoskeleton competition. We performed the sofa, slalom, ramp and door, stepping stones, and stairs tasks in 489 s and 532 s on the first and second races, respectively. We bypassed the tilted path task during both races, as it placed significant strain on our pilot and we did not feel comfortable attempting all of the tasks given our short and aggressive training schedule. We felt that the tasks we completed capitalized on the strengths of our hardware, approach, and pilot, enabling our high placement in the competition.

### Pilot Observations

Our pilot reported high levels of satisfaction with both the Mina v2 hardware and the underlying gait design. He found that the powered ankles assisted in maintaining stability during swing, potentially by having the foot functioning as more a contact patch than a contact point, when compared to exoskeletons with unpowered ankles. This enabled the pilot to better control his center of pressure by shifting his weight throughout the contact surface. This method of balance was found to be much more effective when operating in position-control mode than in impedance-control mode, for much the

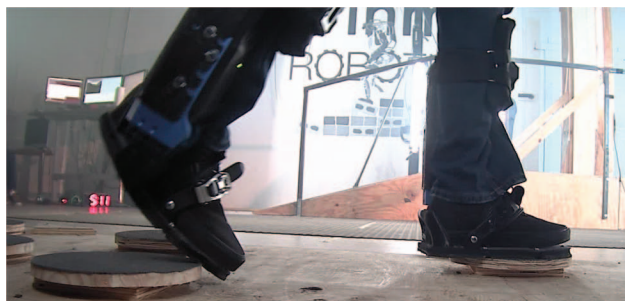

**Figure 8.** An illustration of toe-off motion on stepping stones during training.

72 • IEEE ROBOTICS & AUTOMATION MAGAZINE • DECEMBER 2017

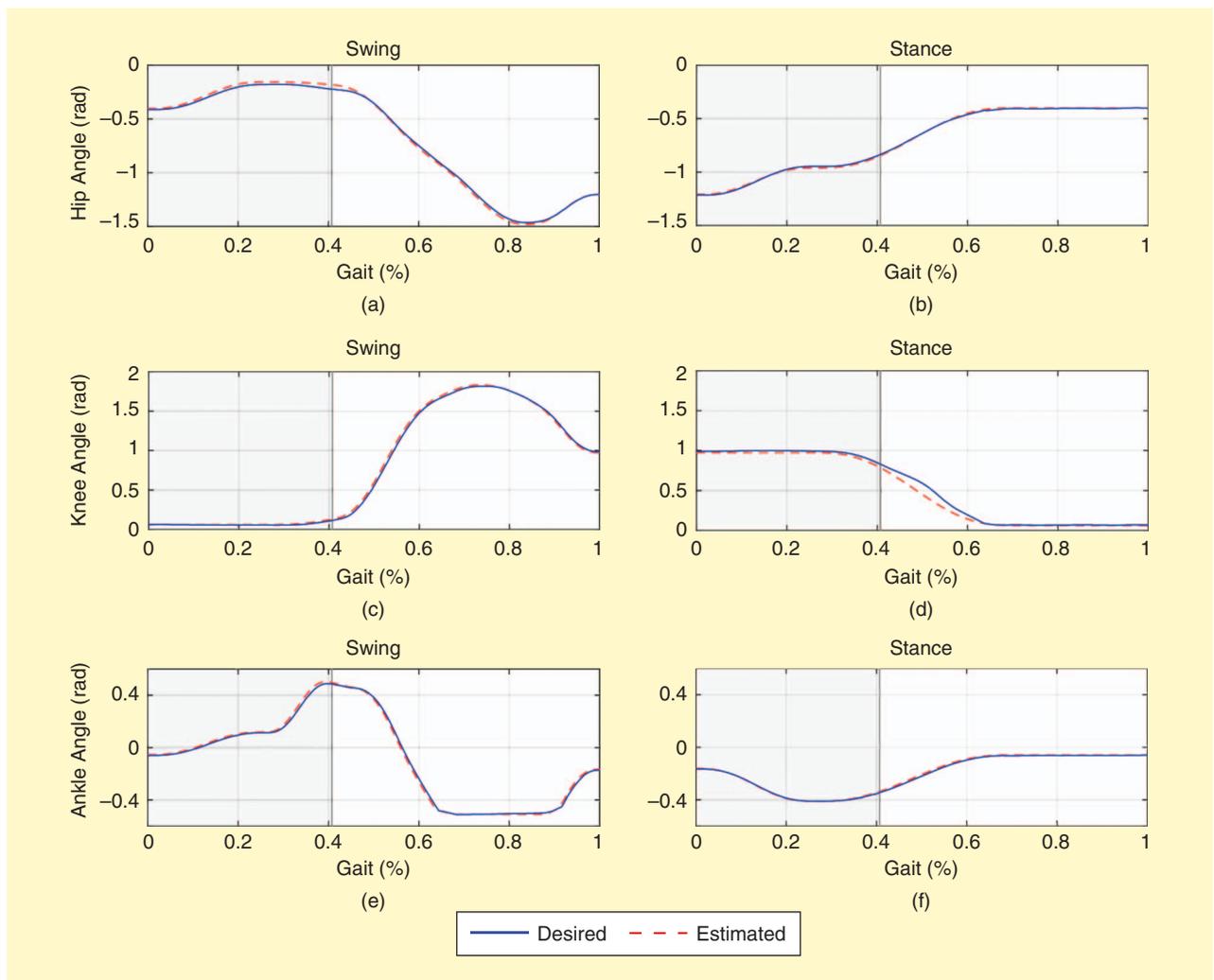

**Figure 9.** The joint angles when ascending stairs with powered ankles; the shaded gray region represents the double-support state: hip angle during (a) swing and (b) stance; knee angle during (c) swing and (d) stance; and ankle angle during (e) swing and (f) stance.

same reasons offered regarding a passive joint. Having tested with Mina v1 and X1, our pilot felt that Mina v2 was the most comfortable exoskeleton he had tried to date, and he found that the inclusion of the toe-off motion assisted greatly in walking. Additionally, he reported that the toe-off motion allowed him to shift more of his weight into the exoskeleton, away from his crutches.

Our pilot took his first step in Mina v2 on 8 August 2016, leaving approximately eight weeks of training time for the competition. Because much of this time was spent tuning, we believe that a new user could be able to competently walk in Mina v2 after only a few days. This is, however, largely a function of the user's timidity, or lack thereof, with our pilot being notably willing to try and fail throughout the training process. We also believe that the inclusion of the powered ankle greatly increased both the pilot's mobility and speed of training, as it decreased his reliance on the crutches and functioned to propel him forward when walking. When coupled with the right attitude, this made for an expedient training process.

## Conclusions

This work presents Team IHMC's entry in the Powered Exoskeleton competition at the 2016 Cybathlon, a competition designed to push the development of new exoskeleton technologies for paraplegics. Mina v2 is a new exoskeleton developed by IHMC that includes powered ankle plantar flexion, a feature not present in most other orthotic exoskeletons. We utilize a custom ball-screw–driven actuator with a modular design on all six joints, allowing easier repair and maintenance. Our distributed control architecture allows high-level trajectories to be generated using an embedded computer in the backpack. Our novel joint kinematic trajectory design centers on the inclusion of a powered toe-off motion. When combined with our unique parameterized swing-leg trajectory, the pilot is able to utilize a powered toe-off motion to help push off the ground and so improve performance. This was used throughout all of the tasks in the competition to start or continue walking, as well as to help power the pilot up the ramp and stairs.

While our team placed an impressive second in the 2016 Cybathlon, exoskeletons with powered ankles offer significantly greater potential. Without powered ankle plantar flexion, true dynamic walking in exoskeletons without crutches is not possible. The inclusion of powered ankles also enables the development of balancing strategies that could potentially remove the need for crutches altogether. This represents a major step forward in exoskeleton research, enabling further development not possible otherwise.

### Acknowledgments

Support for this article comes from National Science Foundation Award IIS–1427213 and from Stormy Dawn Anderson. We also thank our media team, Billy Howell and Jason Conrad, for giving us a fun work environment and for putting out wonderful videos. We also thank the rest of the IHMC Robotics Lab for all their help and support throughout this project. Finally, we would like to thank our pilot, Mark Daniel; without his wonderful attitude, incredible work ethic, and bravery to try new things, none of this would have been possible.

*Robert Griffin,* College of Engineering, Virginia Polytechnic Institute and State University, Blacksburg, and Florida Institute for Human and Machine Cognition, Pensacola. E-mail: rgriffin@ihmc.us.

*Tyson Cobb*, Florida Institute for Human and Machine Cognition, Pensacola. E-mail: tcobb@ihmc.us.

*Travis Craig*, Florida Institute for Human and Machine Cognition, Pensacola. E-mail: tcraig@ihmc.us.

*Mark Daniel*, Florida Institute for Human and Machine Cognition. E-mail: mdaniel@ihmc.us.

*Nick van Dijk*, Florida Institute for Human and Machine Cognition, Pensacola, and Delft University of Technology, The Netherlands. E-mail: nick@gerritsen.tv.

*Jeremy Gines*, Florida Institute for Human and Machine Cognition, Pensacola. E-mail: jgines@ihmc.us.

*Koen Krämer*, Florida Institute for Human and Machine Cognition, Pensacola, and Delft University of Technology, The Netherlands. E-mail: koen.kramer@live.nl.

*Shriya Shah*, College of Engineering, Polytechnic Institute and State University, Blacksburg, and Florida Institute for Human and Machine Cognition, Pensacola. E-mail: sshriya7@vt.edu.

*Olger Siebinga*, Florida Institute for Human and Machine Cognition, Pensacola, and Delft University of Technology, The Netherlands. E-mail: osiebinga@ihmc.us.

*Jesper Smith*, Florida Institute for Human and Machine Cognition, Pensacola. E-mail: jsmith@ihmc.us.

*Peter Neuhaus*, Florida Institute for Human and Machine Cognition, Pensacola. E-mail: pneuhaus@ihmc.us.